\documentclass[runningheads]{llncs}

\usepackage{amssymb}
\usepackage{graphicx}

\usepackage{url}
\urldef{\mailsa}\path|Adrian.Paschke@gmx.de|
\newcommand{\keywords}[1]{\par\addvspace\baselineskip
\noindent\keywordname\enspace\ignorespaces#1}

\begin{document}

\mainmatter\small{  % start of an individual contribution

\title{Verification, Validation and Integrity of Distributed and Interchanged Rule Based Policies and Contracts in the Semantic Web}

\titlerunning{V\&V\&I of Rule Based Policies, Contracts, SLAs}


\author{Adrian Paschke}

\authorrunning{Paschke, Adrian}

% the affiliations are given next
\institute{Internet-based Information Systems, Dept. of Informatics, TU Munich, Germany\\
\mailsa\\}

\toctitle{Verification, Validation and Integrity of Rule Based
Policies and Contracts} \tocauthor{Adrian Paschke} \maketitle

\begin{abstract}
Rule-based policy and contract systems have rarely been studied in
terms of their software engineering properties. This is a serious
omission, because in rule-based policy or contract representation
languages rules are being used as a declarative programming
language to formalize real-world decision logic and create IS
production systems upon. This paper adopts an SE methodology from
extreme programming, namely test driven development, and discusses
how it can be adapted to verification, validation and integrity
testing (V\&V\&I) of policy and contract specifications. Since,
the test-driven approach focuses on the behavioral aspects and the
drawn conclusions instead of the structure of the rule base and
the causes of faults, it is independent of the complexity of the
rule language and the system under test and thus much easier to
use and understand for the rule engineer and the user.

\end{abstract}

\section{Test-driven V\&V for Rule-based Policies and Contracts}

Increasing interest in industry and academia in higher-level
policy and contract languages has led to much recent development.
Different representation approaches have been propose, reaching
from general syntactic XML markup languages such as WS-Policy,
WS-Agreement or WSLA to semantically-rich (ontology based) policy
representation languages such as Rei, KAoS or Ponder and highly
expressive rule based contract languages such as the RBSLA
language \cite{Paschke05RBSLA} or the SweetRules approach. In this
paper we adopt the rule-based view on expressive high-level policy
and contract languages for representing e.g. SLAs, business
policies and other contractual, business-oriented decision logic.
In particular, we focus on logic programming techniques. Logic
programming has been one of the most successful representatives of
declarative programming. It is based on solid and well-understood
theoretical concepts and has been proven to be very useful for
rapid prototyping and describing problems on a high abstraction
level. In particular, the domain of contractual agreements,
high-level policies and business rules' decision logic appears to
be highly suitable to logic programming. For instance, IT service
providers need to manage and possibly interchange large amounts of
SLAs / policies / business rules which describe behavioral,
contractual or business logic using different rule types to
describe e.g. complex conditional decision logic (derivation
rules), reactive or even proactive behavior (ECA rules), normative
statements and legal rules (deontic rules), integrity definitions
(integrity constraints) or defaults, rule preferences and
exceptions (non-monotonic defeasible rules). Such rule types have
been shown to be adequately represented and formalized as logic
programs (LPs) - see the ContractLog KR \cite{Paschke06} developed
in the RBSLA project \cite{PaschkeRBSLA}. However, the rule-based
policy and contract representation imposes some specific needs on
the engineering and life-cycle management of the formalized
specifications: The policy rules must be necessarily modelled
evolutionary, in a close collaboration between domain experts,
rule engineers and practitioners and the statements are not of
static nature and need to be continuously adapted to changing
needs. The future growth of policies or contract specifications,
where rules are often managed in a distributed way and are
interchanged between domain boundaries, will be seriously
obstructed if developers and providers do not firmly face the
problem of quality, predictability, reliability and usability
w.r.t. understandability of the results produced by their
rule-based policy/contract systems and programs. Furthermore, the
derived conclusions and results need to be highly reliable and
traceable to count even in the legal sense. This amounts for
verification, validation and integrity testing (V\&V\&I)
techniques, which are much simpler than the rule based
specifications itself, but nevertheless adequate (expressive
enough) to approximate their intended semantics, determine the
reliability of the produced results, ensure the correct execution
in a target inference environment and safeguard the life cycle of
possibly distributed and unitized rules in rule-based policy
projects which are likely to change frequently.\\
Different approaches and methodologies to V\&V of rule-based
systems have been proposed in the literature such as model
checking, code inspection or structural debugging. Simple
operational debugging approaches which instrument the
policy/contract rules and explore its execution trace place a huge
cognitive load on the user, who needs to analyze each step of the
conclusion process and needs to understand the structure of the
rule system under test. On the other hand, typical heavy-weight
V\&V methodologies in Software Engineering (SE) such as
waterfall-based approaches are often not suitable for rule-based
systems, because they induce high costs of change and do not
facilitate evolutionary modelling of rule-based policies with
collaborations of different roles such as domain experts, system
developers and knowledge engineers. Moreover, they can not check
the dynamic behaviors and the interaction between dynamically
updated and interchanged policies/contracts and target execution
environments at runtime. Model-checking techniques and methods
based e.g. on algebraic-, graph- or Petri-net-based
interpretations are computationally very costly, inapplicable for
expressive policy/contract rule languages and presuppose a deep
understanding of both domains, i.e. of the the testing language /
models and of of the rule language and the rule inferences.
Although test-driven Extreme Programming (XP) techniques and
similar approaches to agile SE have been very successful in recent
years and are widely used among mainstream software developers,
its values, principles and practices have not been transferred
into the rule-based policy and contract representation community
yet. V\&V has been an important area of research in the
expert-system and knowledge engineering community in the mid '80s
to the early '90s manly applying debugging techniques or
transformation approaches into analytical models such as graphs or
algebraic structures. However, to the best of our knowledge nearly
no work has been done in building on these results for V\&V of
rule-based policy/contract specifications and on adopting recent
trends in SE to the domain of policy engineering. In this paper,
we adapt a successful methodology of XP, namely test cases (TCs),
to verify and validate correctness, reliability and adequacy of
rule-based policy and contract specifications. It is well
understood in the SE community that test-driven development
improves the quality and predictability of software releases and
we argue that TCs and integrity constraints (ICs) also have a huge
potential to be a successful tool for declarative V\&V of
rule-based policy and contract systems. TCs in combination with
other SE methodologies such as \textit{test coverage measurement}
which is used to quantify the completeness of TCs as a part of the
feedback loop in the development process and \textit{rule base
refinements} (a.k.a. \textit{refactorings})
\cite{conf/ista/DietrichP05} which optimize the existing rule
code, e.g. remove inconsistencies, redundancy or missing knowledge
without breaking its functionality, qualify for typically
frequently changing requirements and models of rule-based policies
and contracts (e.g. SLAs). Due to their inherent simplicity TCs,
which provide an abstracted black-box view on the rules, better
support different roles which are involved during the engineering
process. In our approach TCs are written homogeneously in the
target programming language, i.e. in the contract/policy rule
language, so that they can be managed, maintained and distributed
together with the policies/contracts. Using TCs and ICs to
represent the constraints which describe the intended semantics of
a policy/contract specification gives policy engineers an
expressive but nevertheless easy to use testing language and makes
policies self validating, to an large extend. In a feedback loop
changing requirements and detected faults (bugs) are translated
into new TCs, and the policy specification is then modified until
the old and the new TCs succeed. This also helps to avoid atrophy
of the rule code and the TCs when the policies are dynamically
changed and extended. During rule interchange in open distributed
environment TCs can be used to ensure correct execution of an
interchanged LP in a target execution environment by validating
the interchanged rules with the attached TCs. The further paper is
structured as follows: In section 2 we review basics in V\&V
research. In section 3 we define syntax and semantics of TCs and
ICs for LP based policy/contract specifications. In section 4 we
introduce a declarative test coverage measure which draws on
inductive logic programming techniques. In section 5 we discuss
TCs for V\&V of rule engines and rule interchange. In section 6 we
describe our reference implementation in the ContractLog KR and
integrate our approach into an existing SE test framework (JUnit)
and a rule markup language (RuleML). In section 7 we discuss
related work and conclude this paper with a discussion of the
test-drive V\&V\&I approach for rule-based policies and contracts.

\section{Basics in Rule-based V\&V Research}
V\&V of rule-based policy/contract specifications is vital to
assure that the LP used to formalize the policy/contract rules
performs the tasks which it was designed for. Accordingly, the
term V\&V is used as a rough synonym for "evaluation and testing".
Both processes guarantee that the LP provides the intended answer,
but also imply other goals such as to assure the security or
maintenance and service of
the rule-based system. There are many definitions of V\&V in the SE literature. In the context of V\&V of rule-based policies/contracts we use the following:\\
\begin{scriptsize}
1. \emph{Verification} ensures the technical correctness of a LP.
Akin to traditional software engineering a distinction between
structurally flawed or logically flawed rule bases can be made
with structural checks for redundancy or relevance and semantic checks for consistency, soundness and completeness.\\
2. As discussed by Gonzales \cite{Gonz00} \emph{validation} should
not be confused with verification. Validation is concerned with
the logical correctness of a rule-based system in a particular
environment/situation and domain. Typically, validation is based
on tests, desirably in the real environment and under real
circumstances, where the rule base is considered as a "black box"
which produces certain outputs (answer to queries) given a set of
input data (assertions represented as facts).\\
\end{scriptsize}
During runtime certain parts of the rule based decision logic
should be static and not subjected to changes or it must be
assured that updates do not change this part of the intended
behavior of the policy/contract. A common way to represent such
constraints are ICs. Roughly, if validation is interpreted as:
"\textit{Are we building the right product?}" and verification as:
"\textit{Are we building the product right?}" then integrity might
be loosely defined as: "\textit{Are we keeping the product
right}?", leading to the new pattern: \textbf{V\&V\&I}. Hence, ICs
are a way to formulate consistency (or inconsistency) criteria of
a dynamically updated knowledge base (KB). Another distinction
which can be made is between errors and
anomalies:\\
\begin{scriptsize}
- \emph{Errors} represent problems which directly effect the
operations of a rule base. The simplest source of errors are
typographical mistakes which can be solved by a verifying parser.
More complex problems arise in case of large rule bases
incorporating several people during design and maintenance and in
case of the dynamic alteration of the rule base via adding,
changing or refining the knowledge which might easily
lead to incompleteness and contradictions.\\
- \emph{Anomalies} are considered as symptoms of genuine errors,
i.e. they man not necessarily represent problems in themselves.\\
\end{scriptsize}
Much work has been done to establish and classify the nature of
errors and anomalies that may be present in rule bases, see e.g.
the taxonomy of anomalies from Preece and Shinghal
\cite{PreShi94}. A general distinction can be made between
errors/anomalies concerned with the design of rule bases and those
concerned with the inferences. Typical inference errors are e.g.
redundant rules, circular rules or dead end rules (in forward
chaining systems). Typical design errors/anomalies are e.g.
duplication, inconsistency or subsumedness. For a detailed
discussion of potential errors and anomalies that may occur in
rule bases see e.g. \cite{CoBeCa93}. Here, we briefly review the
notions that are commonly used in the literature
\cite{Plant95,AHPV98,Preece01}, which range from semantic checks
for consistency and completeness to structural checks for
redundancy, relevance and
reachability:\\
\begin{scriptsize}
1. \emph{Consistency}: No conflicting conclusions can be made from
a set of valid input data. The common definition of consistency is
that two rules or inferences are inconsistent if they succeed at
the same knowledge state, but have conflicting results. Several
special cases of
inconsistent rules are considered in literature such as:\\
- \textit{self-contradicting rules} and \textit{self-contradicting rule chains}, e.g.  $p \wedge q \rightarrow \neg p$\\
- \textit{contradicting rules} and \textit{contradicting rule
chains}, e.g. $p \wedge q \rightarrow s$
and $p \wedge q \rightarrow \neg s$\\
Note that the first two cases of  self-contradiction are not
consistent in a semantic sense and can equally be seen as
redundant rules, since they can be never concluded.\\
2. \emph{Correctness/Soundness}: No invalid conclusions can be
inferred from valid input data, i.e. a rule base is correct when
it holds for any complete model $M$, that the inferred output from
valid inputs via the rule base are true in $M$. This is closely
related to \textit{soundness} which checks that the intended
outputs indeed follows from the valid input. Note, that in case of
partial models with only partial information this means that all
possible partial models need to be verified instead of only the
complete models. However, for monotonic inferences these notions
coincide and a rule base which is
sound is also consistent.\\
3. \emph{Completeness:} No valid input information fails to
produce the intended output conclusions, i.e. completeness relates
to gaps (incomplete knowledge) in the knowledge base. The
iterative process of building large rule bases where rules are
tested, added, changed and refined obviously can leave gaps such
as missing rules in the knowledge base. This usually results in
intended derivations which are not possible. Typical sources of
incompleteness are missing facts or rules which prevent intended
conclusions to be drawn. But there are also other sources. A KB
having too many rules and too many input facts negatively
influences performance and may lead to incompleteness due to
termination problems or memory overflows. Hence, superfluous rules
and non-terminating rule chains can be also
considered as completeness problems, e.g.:\\
- \textit{Unused rules and facts}, which are never used in any
rule/query derivation (backward reasoning) or which are
unreachable or dead-ends (forward reasoning).\\
- \textit{Redundant rules} such as identical rules or rule chains,
e.g. $p \rightarrow q$ and $p \rightarrow q$.\\
- \textit{Subsumed rules}, a special case of redundant rules,
where two rules have the same rule head but one rule contains more
prerequisites (conditions) in the body, e.g. $p \wedge q
\rightarrow r$ and $p \rightarrow r$.\\
- \textit{Self-contradicting rules}, such as $p \wedge q \wedge
\neg p \rightarrow r$ or simply $p \rightarrow \neg p$, which can
never
succeed.\\
- \textit{Loops} in rules of rule chains, e.g. $p \wedge q
\rightarrow q$ or tautologies such as $p \rightarrow p$.\\
\end{scriptsize}

\section{Homogeneous Integration of Test Cases and Integrity Constraints into Logic Programs}
The relevance of V\&V of rule bases and LPs has been recognized in
the past (see section 2 and 7) and most recently also in the
context of policy explanations \cite{BOP06}. The majority of these
approaches rely on debugging the derivation trees and giving
explanations (e.g. via spy and trace commands) or transforming the
program into other more abstract representation structures such as
graphs, petri nets or algebraic structures which are then analyzed
for inconsistencies. Typically, the definition of an
inconsistency, error or anomaly (see section 2) is then given in
the language used for analyzing the LP, i.e. the V\&V information
is not expressed in the same representation language as the rules.
This is in strong contrast to the way people would like to
engineer, manage and maintain rule-based policies and systems.
Different skills for writing LPs and analyzing them are needed as
well as different systems for reasoning with rules and for V\&V.
Moreover, the used V\&V methodologies (e.g. model checking or
graph theory) are typically much more complicated than the
rule-based programs. In fact, it turns out that even writing
rule-based systems that are useful in practice is already of
significant complexity, e.g. due to non-monotonic features or
different negations, and that simple methods are needed to
safeguard the engineering and maintenance process w.r.t. V\&V\&I.
Therefore, what policy engineers and practitioners would like to
have is an "easy-to-use" approach that allows representing rules
and tests in the same homogeneous representation language, so that
they can be engineered, executed, maintained and interchanged
together using the same underlying syntax, semantics,
methodologies and execution/inference environment. In this section
we elaborate on this homogeneous integration approach based on the
common "denominator": \textit{extended logic programming}.

In the following we use the standard LP notation with an ISO
Prolog related scripting syntax called Prova \cite{KPS06} and we
assume that the reader is familiar with logic programming
techniques \cite{Llyod87}. For the semantics of the knowledge base
we adapt a rather general definition \cite{DIX95} of LP semantics
which also possibly include some form of non-monotonic reasoning,
because our test-driven approach is intended to be general and
applicable to several logic classes / rule languages (e.g.
propositional, DataLog, normal, extended) in order to fulfill the
different KR needs of particular policy or contract representation
projects (e.g. w.r.t expressiveness and computational complexity
which are in a trade-off relation to each other). In particular,
as we will show in section 5, TCs can be also used to verify the
possible unknown semantics of a target inference service in a open
environment such as the (Semantic) Web and test the correct
execution of an interchanged policy/contract in the target
environment.\\
\textit{
\begin{scriptsize}
    - A semantics $SEM(P)$ of a LP $P$ is proof-theoretically
    defined as a set of literals that are derivable from $P$ using a
    particular derivation mechanisms, such as linear
    SLD(NF)-resolution variants with negation-as-finite-failure rule or non-linear tabling approaches such as
    SLG resolution. Model-theoretically, a semantics
    $SEM(P)$ of a program $P$ is a subset of all models of $P$:
    $MOD(P)$. In this paper in most cases a subset of the (3-valued)
    Herbrand-models of the language of $L_{P}$: $SEM(P) \subseteq
    MOD_{Herb^{L_{P}}} (P)$. Associated to SEM(P) are two entailment relations:\\
    1. sceptical, where the set of all atoms or default atoms are
    true in all models of SEM(P)\\
    2. credulous, where the set of all atoms or default atoms are
    true in at least one model of SEM(P)
    - A semantics $SEM'$
    extends a semantics $SEM$ denoted by $SEM' \geq SEM$, if for all
    programs $P$ and all atoms $l$ the following holds: $SEM(P)
    \models l \Rightarrow SEM'(P) \models l$, i.e. all atoms derivable
    from $SEM$ with respect to P are also derivable from $SEM'$, but
    $SEM'$ derives more true or false atoms than $SEM$. The semantics $SEM'$ is defined for a class of programs that strictly includes the class of programs with the semantics $SEM$.
    $SEM'$ coincides with $SEM$ for all programs of the class of
    programs for which $SEM$ is defined.
\end{scriptsize}}

In our ContractLog reference implementation we mainly adopt the
sceptical viewpoint on extended LPs and apply an extended linear
SLDNF variant as procedural semantics which has been extended with
explicit negation, goal memoization and loop prevention to
overcome typical restrictions of standard SLDNF and compute WFS
(see ContractLog inference engine).

The general idea of TCs in SE is to predefine the intended output
of a program or method and compare the intended results with the
derived results. If both match, the TC is said to capture the
intended behavior of the program/method. Although there is no
100\% guarantee that the TCs defined for V\&V of a program exclude
every unintended results of the program, they are an easy way to
approximate correctness and other SE-related quality goals (in
particular when the TCs and the program are refined in an
evolutionary, iterative process with a feedback loop). In logic
programming we think of a LP as formalizing our knowledge about
the world and how the world behaves. The world is defined by a set
of models. The rules in the LP constrain the set of possible
models to the set of models which satisfy the rules w.r.t the
current knowledge base (actual knowledge state). A query $Q$ to
the LP is typically a conjunction of literals (positive or
negative atoms) $G_{1} \wedge .. \wedge G_{n}$, where the literals
$G_{i}$ may contain variables. Asking a query $Q$ to the LP then
means asking for all possible substitutions $\theta$ of the
variables in $Q$ such that $Q \theta$ logically follows from the
LP $P$ and $P \models Q$. The substitution set $\theta$ is said to
be the answer to the query, i.e. it is the output of the program
$P$. Hence, following the idea of TCs, for V\&V of a LP $P$ we
need to predefine the intended outputs of $P$ as a set of (test)
queries to $P$ and compare it with the actual results / answers
derived from $P$ by asking these test queries to $P$. Obviously,
the set of possible models of a program might be quite large (even
if many constraining rules exist), e.g. because of a large fact
base or infinite functions. As a result the set of test queries
needed to test the program and V\&V of the actual models of $P$
would be in worst case also infinite. However, we claim that most
of the time correctness of a set of rules can be assured by
testing a much smaller subset of these models. In particular, as
we will see in the next section, in order to be an adequate cover
for a LP the tests need to be only a least general instantiation
(specialization) of the rules' terms (arguments) in order to fully
investigate and test all rules in $P$. This also supports our
second claim, that V\&V of LPs with TC can be almost ever done in
reasonable time, due to the fact that the typical test query is a
ground query (without variables) which has a small search space
(as compared to queries with free variables) and only proves
existence of at least one model
satisfying it. In analogy to TCs in SE we define a TC as $TC:=\{A,T\}$ for a LP $P$ to consists of: \\
\begin{scriptsize}
1. a set of possibly empty input \textit{assertions} "$A$" being
the set of temporarily asserted test input facts (and additionally
meta test rules - see section 5) defined over the alphabet "$L$".
The assertions are used to temporarily setup the test environment.
They can be e.g. used to define test facts, result values of
(external) functions called by procedural attachments, events and
actions for testing reactive rules or additional meta test rules.\\
2. a set of one ore more \textit{tests} §$T$§. Each test $T_{i}$, $i>0$  consists of:\\
- a \textit{test query} $Q$ with goal literals of the form
$q(t_{1},..t_{n})?$, where $Q \in rule(P)$ and $rule(P)$ is the set of literals in the head of rules (since only rules need to be tested)\\
- a \textit{result} $R$ being either a positive "$true$", negative "$false$" or "$unknown$" label.\\
- an intended \textit{answer set} $\theta$ of expected variable
bindings for the variables of the test query $Q$:
$\theta:=\{X_{1},..X_{n}\}$ where each $X_{i}$ is a set of
variable bindings $\{X_{i}/a_{1},..,X_{i}/a_{n}\}$. For ground
test queries $\theta:=\emptyset$.\\
\end{scriptsize}
We write a TC $T$ as follows: $T=A \cup \{Q=>R:\theta\}$. If a TC
has no assertions we simply write $T=\{Q=>R:\theta\}$. For
instance, a TC $T1=\{p(X)=>true
 :\{X/a,X/b,X/b\}, q(Y)=>false\}$ defines a TC $T1$ with two test
 queries $p(X)?$ and $q(Y)?$. The query $p(X)?$ should succeed and return three answers $a$,$b$ and
 $c$ for the free variable $X$. The query $q(Y)$ should fail. In case we are only interested in the existential success of a test query we shorten the notation of a TC to
$T=\{Q=>R\}$.

To formulate runtime consistency criteria w.r.t. conflicts which might arise due to knowledge updates, e.g. adding rules, we apply ICs:\\
\begin{scriptsize}
\textit{ An IC on a LP is defined as a set of conditions that the
constrained KB must satisfy, in order to be considered as a
consistent model of the intended (real-world domain-specific)
model. \emph{Satisfaction} of an IC is the fulfillment to the
conditions imposed by the constraint and \emph{violation} of an IC
is the fact of not giving strict fulfillment to the conditions
imposed by the constraint, i.e. satisfaction resp. violation on a
program (LP) $P$ w.r.t the set of $IC:=\{ic_{1},..ic_{i}\}$
defined in $P$ is the satisfaction of each $ic_{i} \in IC$ at each
KB state $P':= P \cup M_{i} \Rightarrow P \cup M_{i+1}$ with
$M_{0}= \emptyset$, where $M_{i}$ is an arbitrary knowledge update
adding,removing or changing rules or facts to the dynamically extended or reduced KB.}\\
\end{scriptsize}
Accordingly, ICs are closely related to our notion of TCs for LPs.
In fact, TCs can be seen as more expressive ICs. From a
syntactical perspective we distinguish ICs from TCs, since in our
(ContractLog) approach we typically represent and manage TCs as
stand-alone LP scripts (module files) which are imported to the
KB, whereas ICs are defined as LP functions. Both, internal ICs or
external TCs can be used to define conditions which denote a logic
or application specific conflict. ICs in ContractLog are defined
as a n-ary function $integrity(<operator>,<conditions>)$. We
distinguish four types of ICs:\\
\begin{scriptsize}
    - \emph{Not-constraints} which express that none of the stated conclusions should be drawn.\\
    - \emph{Xor-constraints} which express that the stated conclusions should not be drawn at the same time.\\
    - \emph{Or-constraints} which express that at least one of the stated conclusions must be drawn.\\
    - \emph{And-constraints} which express that all of the stated conclusion must draw.\\
\end{scriptsize}
ICs are defined as constraints on the set of possible models and
therefore describe the model(s) which should be considered as
strictly conflicting. Model theoretically we attribute a 2-valued
truth value (true/false) to an IC and use the defined set of
constraints (literals) in an IC as a goal on the program $P$, by
meta interpretation (as procedural semantics) of the integrity
functions. In short, the truth of an IC in a finite interpretation
$I$ is determined by running the goal $G_{IC}$ defined by the IC
on the clauses in $P$ or more precisely on the actual knowledge
state of $P_{i}$. If the $G_{IC}$ is satisfied, i.e. there exists
at least one model for the sentence formed by the $G_{IC}$: $P_{i}
\models G_{IC}$, the IC is violated and $P$ is proven to be in an
inconsistent state w.r.t. $IC$: $IC$ is violated resp. $P_{i}$
violates integrity iff for any interpretation $I$, $I \models
P_{i} \rightarrow I \models G_{IC}$. We define the following
interpretation for ICs:\\
\begin{scriptsize}
And $and(C_{1},..,C_{n})$: $P_{i} \models (not C_{1}
\vee .. \vee not C_{n})$ if exists $i \in {1,..,n}, P_{i} \models not$ $C_{i}$\\
Not: $not(C_{1},..,C_{n})$: $P_{i} \models (C_{1} \vee
.. \vee C_{n})$ if exists $i \in {1,..,n}, P_{i} \models C_{i}$\\
Or: $or(C_{1},..,C_{n})$: $P_{i} \models (not C_{1} \wedge
.. \wedge not C_{n}$ if for all $i \in {1,..,n}, P_{i} \models not$ $C_{i}$\\
Xor: $xor(C_{1},..,C_{n})$: $P_{i} \models (C_{j} \wedge
 C_{k})$ if exists $j \in {1,..,n}, P_{i} \models C_{j}$ and exists $k \in {1,..,n}, P_{i} \models C_{k}$  with $C_{j} \neq C_{k}$ and $C_{j} \in C$, $C_{k} \in C$\\
\end{scriptsize}
$C:=\{C_{1},..C_{n}\}$ are positive or negative (explicit negated)
n-ary atoms which might contain variables; $not$ is used in the
usual sense of default negation, i.e. if a constraint literal can
not be proven true, it is assumed to be false. If there exists a
model for a IC goal (as defined above), i.e. the "integrity test
goal" is satisfied $P_{i} \models G_{IC}$, the IC is assigned true
and hence integrity is violated in the actual knowledge/program
state $P_{i}$.

\section{Declarative Test Coverage Measurement}
Test coverage is an essential part of the feedback loop in the
test-driven engineering process. The coverage feedback highlights
aspects of the formalized policy/contract specification which may
not be adequately tested and which require additional testing.
This loop will continue until coverage of the intended models of
the formalized policy specification meets an adequate
approximation level by the TC resp. test suites (TS) which bundle
several TCs. Moreover, test coverage measurements helps to avoid
atrophy of TSs when the rule-based specifications are evolutionary
extended. Measuring coverage helps to keep the tests up to a
required level if new rules are added or existing rules are
removed/changed. However, conventional testing methods for
imperative programming languages rely on the control flow graph as
an abstract model of the program or the explicitly defined data
flow and use coverage measures such as branch or path coverage. In
contrast, the proof-theoretic semantics of LPs is based on
resolution with unification and backtracking, where no explicit
control flow exists and goals are used in a refutation attempt to
specialize the rules in the declarative LP by unifying them with
the rule heads. Accordingly, building upon this central concept of
unification \emph{a test covers a logic program $P$, if the test
queries (goals) lead to a least general specialization of each
rule in $P$, such that the full scope of terms (arguments) of each literal in each rule is investigated by the set of test queries.}\\
Inductively deriving general information from specific knowledge
is a task which is approached by inductive logic programming (ILP)
techniques which allow computing the least general generalization
(lgg), i.e. the most specific clause (e.g. w.r.t. theta
subsumption) covering two input clauses. A lgg is the
generalization that keeps an generalized term $t$ (or clause) as
special as possible so that every other generalization would
increase the number of possible instances of $t$ in comparison to
the possible instances of the lgg. Efficient algorithms based on
syntactical anti-unification with $\theta$-subsumption ordering
for the computation of the (relative) lgg(s) exist and several
implementations have been proposed in ILP systems such as GOLEM,
or FOIL. $\theta$-subsumption introduces a syntactic notion of
generality: A rule (clause) $r$ (resp. a term $t$)
$\theta$-subsumes another rule $r'$, if there exists a
substitution $\theta$, such that $r \subseteq r'$, i.e. a rule $r$
is \textit{as least as general as} the rule $r'$ ($r \leq r'$), if
$r$ $\theta$-subsumes $r'$ resp. \textit{is more general than}
$r'$ ($r < r'$) if $r \leq r'$ and $r'\nleq r$. (see e.g.
\cite{PLOTKIN70}) In order to determine the level of coverage the
specializations of the rules in  the LP under test are computed
via specializing the rules with the test queries by standard
unification. Then via generalizing these specializations under
$\theta$-subsumption ordering, i.e. computing the lggs of all
successful specializations, a reconstruction of the original LP is
attempted. The number of successful "recoverings" then give the
level of test coverage, i.e. the level determines those statements
(rules) in a LP that have been executed/investigated through a
test run and those which have not. In particular, if the complete
LP can be reconstructed via generalization of the specialization
then the test fully covers the LP. Formally we express this as follows:\\
Let $T$ be a test with a set of test queries
$T:=\{Q_{1}?,..,Q_{n}?\}$ for a program $P$, then $T$ is a cover
for a rule $r_{i} \in P$, if the $lgg(r_{i}') \simeq r_{i}$ under
$\theta-subsumption$, where $\simeq$ is an equivalence relation
denoting variants of clauses/terms and the $r_{i}'$ are the
specializations of $r_{i}$ by a query $Q_{i} \in T$. It is a cover
for a program $P$, if $T$ is a cover for each rule $r_{i} \in P$.
With this definition it can be determined whether a test covers a
LP or not. The coverage measure for a LP $P$ is then given by the
number of covered rules $r_{i}$ divided by the number $k$ of all
rules in $P$:\\
\\
$cover_{P}(T):- \frac{\sum_{i=1}^{k} cover_{r_{i}}(T)}{k}$\\

For instance, consider the following simplified business policy
$P$: {\scriptsize
\begin{verbatim}
discount(Customer, 10%) :- gold(Customer).
gold(Customer) :- spending(Customer, Value) , Value > 3000.
spending('Moor',5000). spending('Do',4000). %facts
\end{verbatim}
Let $T =\{discount('Moor',10\%)?=>true,
discount('Do',10\%)?=>true$ be a test with two test queries. The
set of directly derived specializations by applying this tests on
$P$ are:
\begin{verbatim}
discount('Moor',10%) :- gold('Moor').
discount('Do',10%) :- gold('Do').
\end{verbatim}
The computed lggs of this specializations are:
\begin{verbatim}
discount(Customer,10%) :- gold(Customer).
\end{verbatim}
Accordingly, the coverage of $P$ is $50\%$. We extend $T$ with the
additional test goals: \{$gold('Moor')?=>true,
gold('Do')?=>true)?$\}. This leads to two new specializations:
\begin{verbatim}
gold('Moor') :- spending('Moor',Value) , Value > 3000. gold('Do')
:- spending('Do',Value) , Value > 3000.
\end{verbatim}
The additional lggs are then:
\begin{verbatim}
gold(Customer) :- spending(Customer, Value) , Value > 3000.
\end{verbatim}
$T$ now covers $P$, i.e. coverage = $100\%$.}\\
\\
The coverage measure determines how much of the information
represented by the rules is already investigated by the actual
tests. The actual lggs give feedback how to extend the set of test
goals in order to increase the coverage level. Moreover,
repeatedly measuring the test coverage each time when the rule
base becomes updated (e.g. when new rules are added) keeps the
test suites (set of TCs) up to acceptable testing standards and
one can be confident that there will be only minimal problems
during runtime of the LP because the rules do not only pass their
tests but they are also well tested. In contrast to other
computations of the lggs such as implication (i.e. a stronger
ordering relationship), which becomes undecidable if functions are
used, $\theta$-subsumption has nice computational properties and
it works for simple terms as well as for complex terms with or
without negation, e.g. $p() :- q(f(a))$ is a specialization of $p
:- q(X)$. Although it must be noted that the resulting clause
under generalization with $\theta$-subsumption ordering may turn
out to be redundant, i.e. it is possible find an equivalent one
which is described more shortly, this redundancy can be reduced
and since we are only generalizing the specializations on the top
level this reduction is computationally adequate. Thus,
$\theta$-subsumption and least general generalization qualify to
be the right framework of generality in the application of our
test coverage notion.

Although, the defined coverage measure is based on the central
concept of unification and uses ILP techniques for generalization
of the derived specializations of the rule base, it is worth
noting, that the measure might be applied also in the context of
forward-directed reactive rules such as ECA rules or production
rules. There are several approaches in the active database domain
which transform active rules into LP derivation rules, in order to
exploit the formal declarative semantics of logic programs to
overcome confluence and termination problems of active rule
execution sequences, where the actions are input events of further
active rules. \cite{Zan95,FlGr98,BaLo96} For such transformed
declarative rule bases consisting of LP derivation rules test
cases can be written and the coverage can be computed as described
above. The combination of deductive and active rules has been also
investigated in different approaches mainly based on the
simulation of active rules by means of deductive rules.
\cite{Luda98,LaLuMa98,Zan93} Moreover, there are approaches which
directly build reactive rules on top of LP derivation rules such
as the Event Condition Action Logic Programming language (ECA-LP)
which enables a homogeneous representation of ECA rules and
derivation rules. \cite{ECA-RuleML,ECA-LP} Closely related are
also logical update languages such as transaction logics and in
particular serial Horn programs, where the serial Horn rule body
is a sequential execution of actions in combination with standard
Horn pre-/post conditions. \cite{BonKi95} These serial rules can
be processed top-down or bottom-up and hence are closely related
to the production rules style of $condition \rightarrow update$
$action$. This partial relation between backward reasoning LP
derivation rules and forward reasoning production rules which
enables transformations of production rule bases into logic
programs has been also shown for a subclass of production rules,
the stratified production rules. Hence, these class of production
rules also qualifies for our goal-driven testing approach and
unification based test coverage measure. \cite{DuMan02,Rasch94}

\section{Test-driven V\&V of Rule Engines and Rule Interchange}
Typical rule-based B2B contracts or service-oriented policies are
managed and maintained in a distributed environment where the
rules and data is interchanged over domain and system boundaries
using more or less standardized rule markup interchange formats,
e.g. RuleML, SWRL, RBSLA, RIF. The interchanged rules/LPs need to
be interpreted and correctly executed in the target environment,
i.e. in a target rule/inference engine, which might be provided as
an open (Web) service by a third-party provider or a
standardization body such as OMG or W3C (see \cite{PaDiBo05}).
Obviously, the correct execution of the interchanged LP depends on
the semantics of both, the LP and the the inference engine (IE).
TCs, which are interchanged together with the LP, can be used to
test whether the LP still behaves as intended in the target
environment.

To address this issues the IE, the interchanged LP and the
provided TCs must reveal their (intended resp. implemented)
semantics. This might be solved with explicit meta annotations
based on a common vocabulary, e.g. an (Semantic Web) ontology
which classifies semantics such as \textit{COMP} (completion
semantics), \textit{STABLE} (stable model), \textit{WFS}
(well-founded) and relates them to classes of LPs such as
\textit{positive definite LPs}, \textit{stratified LPs},
\textit{normal LPs}, \textit{extended LPs}, \textit{disjunctive
LPs}. The ontology can then be used to describe additional meta
information about the semantics and logic class of the
interchanged rules and TCs and find appropriate IEs to correctly
and efficiently interpret and execute the LP, e.g. (1) via
configuring the rule engine for a particular semantics in case it
supports different ones (see e.g. the configurable ContractLog
IE), (2) by executing an applicable variant of several
interchanged semantics alternatives of the LP or (3) by automatic
transformation approaches which transform the interchange LP into
an executable LP. However, we do not believe that each rule engine
vendor will annotate its implementation with such meta
information, even when there is an official standard Semantic Web
ontology on hand (e.g. released by OMG or W3C). Therefore, means
to automatically determine the supported semantics of IEs or LPs
are needed. As we will show, TCs can be extended to \emph{meta
test programs} testing typical properties of well-known semantics
and by the combination of succeed and failed meta tests uniquely
determine the unknown semantics of the target environment.

A great variety of semantics for LPs (LP-semantics) and
non-monotonic reasoning (NMR-semantics) have been developed in the
past decades. For an overview we relate to \cite{DIX95}. In
general, there are three ways to determine the semantics (and
hence the IE) to be used for execution: (1) by its
\emph{complexity and expressiveness class} (which are in a
trade-off relation to each other), (2) by its \emph{runtime
performance} or (3) by the \emph{semantic properties} it should
satisfy.  A generally accepted criteria as to why one semantics
should be used over another does not exists, but two main
competing approaches, namely WFS and STABLE, have been broadly
accepted as declarative semantics for normal LPs.

For discussion of the worst case complexity and expressiveness of
several classes of LPs we refer to \cite{DEGV97}. Based on these
worst-case complexity results for different semantics and
expressive classes of LPs, which might be published in a machine
interpretable format (Semantic Web ontology) for automatic
decision making, certain semantics might be already excluded to be
usable for a particular rule-based policy/contract application.
However, asymptotic worst-case results are not always appropriate
to quantify performance and scalability of a particular rule
execution environment since implementation specifics of an IE such
as the use of inefficient recursions or memory-structures might
lead to low performance or memory overflows in practice. TCs can
be used to measure the runtime performance and scalability for
different outcomes of a rule set given a certain test fact base as
input. By this certain points of attention, e.g., long
computations, loops or deeply nested derivation trees, can be
identified and a refactoring of the rule code (e.g. reordering
rules, narrowing rules, deleting rules etc.) can be attempted
\cite{conf/ista/DietrichP05}. We call this \emph{dynamic testing}
in opposite to \emph{functional testing}. \emph{Dynamic TCs} with
maximum time values (time constraints) are defined as an extension
to functional TCs (see section 3): $TC=A \cup \{Q=>R: \theta
<MS\}$, where MS is a maximum time constraint for the test query
$Q$. If the query was not successful within this time frame the
test is said to be failed. For instance, consider the dynamic TC
$TC_{dyn}: {q(a)?=>true < 1000 ms}$. The test succeeds iff the
test query succeeds and the answer is computed in less than 1000
milliseconds.

To define a meta ontology of semantics and LP classes (represented
as a OWL ontology - see \cite{PaVV05} for more details) which can
be used to meta annotate the interchanged policy LPs, the IEs and
the TCs we draw on the general semantics classification theory
developed by J. Dix \cite{Dix95a,Dix95b}. Typical top-level LP
classes are, e.g., definite LPs, stratified LPs, normal LP,
extended LPs, disjunctive LPs. Well-known semantics for these
classes are e.g., least and supported Herbrand models, 2 and
3-valued COMP, WFS, STABLE, generalized WFS etc. Given the
information to which class a particular LP belongs or which is the
intended semantics of this LP and given the information which
semantics is implemented by the (target) IE, it is straightforward
to decide wether the LP can be executed by the IE or not at all.
In short, a LP can not be executed by an IE, if the IE derives
less literals than the intended $SEM$ for which the LP was design
for would do, i.e. $SEM'(IE) \geq SEM(P)$ or the semantics
implemented by the IE is not adequate for the program, i.e.
$SEM'(IE) \neq SEM(P)$ . This information can be give by meta
annotations, e.g., \emph{class:} defines the class of the LP / IE;
\emph{semantics:} defines the semantics of the LP / IE;
\emph{syntax:} defines the rule language syntax.

In the context of rule interchange with open, distributed IEs,
which might be provided as public services, an important question
is, wether the IE correctly implements a semantics. \emph{Meta
TCs} can be used for V\&V of the interchanged LP in the target
environment and therefore establish trust to this service.
Moreover, meta TCs checking general properties of semantics can be
also used to verify and determine the semantics of the target IE
in case it is unknown (not given by meta annotations). Kraus et
al. \cite{KLM90} and Dix \cite{Dix95a,Dix95b} proposed several
weak and structural (strong) properties for arbitrary
(non-monotonic) semantics, e.g.:\\

\begin{scriptsize}
\textbf{Strong Properties}\\
- \emph{Cumulativity}: If $U \subseteq V \subseteq
    SEM^{scept}_{P}(U)$, then $SEM^{scept}_{P}(U) =
    SEM^{scept}_{P}(V)$, where $U$ and $V$ are are sets of atoms
    and $SEM^{scept}_{P}$ is an arbitrary sceptical semantics for
    the program $P$, i.e. if $a \mid\sim b$ then $a \mid\sim c$ iff $(a \wedge b) \mid\sim
    c$.\\
- \emph{Rationality}: If $U \subseteq V, V \cap
\{A:SEM^{scept}_{P}(U) \models
    \neg A\} = \emptyset$, then $SEM^{scept}_{P}(U) \subseteq
    SEM^{scept}_{P}(V)$.\\
\\
\textbf{Weak Properties }\\
- \emph{Elimination of Tautologies}: If a rule $a \leftarrow b
    \wedge not$ $c$ with $a \cap b = \emptyset$ is eliminated from
    a program $P$, then the resulting program $P'$ is semantically
    equivalent: $SEM(P) = SEM(P')$. $a$,$b$,$c$ are sets of
    atoms: $P \mapsto P'$ iff there is a rule $H \leftarrow B \in
    P$ such that $H \in B$ and $P'=P \setminus \{H \leftarrow
    b\}$\\
- \emph{Generalized Principle of Partial Evaluation (GPPE)}: If a
    rule $a \leftarrow b \wedge not$ $c$, where $b$ contains an
    atom $B$, is replaced in a program $P'$ by the $n$ rules $a
    \cup (a^{i}-{B}) \leftarrow ((b-{B}) \cup b^{i}) \wedge not$
    $(c \cup c^{i})$, where $a^{i} \leftarrow b^{i} \wedge not$
    $c^{i} (i=1,..n)$ are all rules for which $B \in a^{i}$, then
    $SEM(P) = SEM(P')$\\
- \emph{Positive/Negative Reduction}: If a rule $a \leftarrow b
    \wedge not$ $c$ is replaced in a program $P'$ by $a \leftarrow
    b \wedge not$ $(c-C)$ (C is an atom), where $C$ appears in no
    rule head, or a rule $a \leftarrow b \wedge not$ $c$ is deleted from
    $P$, if there is a fact $a'$ in $P$ such that $a' \subseteq
    c$, then $SEM(P) = SEM(P')$:\\
    1. Positive Reduction: $P \mapsto P'$ iff there is a rule $H \leftarrow B \in
    P$ and a negative literal $not$ $B \in B$ such that $B \ni
    HEAD(P)$ and $P'=(P \setminus \{H \leftarrow B\}) \cup \{H
    \leftarrow (B \setminus \{notB\})\}$\\
    2. Negative Reduction: $P \mapsto P'$ iff there is a rule $H \leftarrow B \in
    P$ and a negative literal $not$ $B \in B$ such that $B \in FACT(P)$ and $P'=(P \setminus \{H \leftarrow
    B\})$\\
- \emph{Elimination of Non-Minimal Rules / Subsumption}: If a rule
$a \leftarrow b \wedge not$ $c$ is deleted from a program $P$ if
    there is another rule $a' \leftarrow b' \wedge not$ $c'$ such
    that $a' \subseteq a, b' \subseteq b, c' \subseteq c$, where
    at least one $\subseteq$ is proper, then $SEM(P) = SEM(P')$: $P \mapsto P'$ iff there are rules $H \leftarrow B$ and $H \leftarrow B' \in
    P$ such that $B \subset B'$ and $P'=P \setminus \{H \leftarrow
    B'\}$\\
- \emph{Consistency}: $SEM(P)=\emptyset$ for all disjunctive LPs\\
- \emph{Independence}: For every literal $L$, $L$ is true in
    every $M \in SEM(P)$ iff $L$ is true in every $M \in SEM(P \cup P')$
    provided that the language of $P$ and $P'$ are disjoint and
    $L$ belongs to the language of $P$\\
- \emph{Relevance}: The truth value of a literal $L$ with
    respect to a semantics $SEM(P)$, only depends on the
    subprogram formed from the \textit{relevant rules} of $P$ ($relevant(P)$) with respect
    to $L$: $SEM(P)(L)=SEM(relevant(P,L))(L)$
\end{scriptsize}\\

The basic idea to apply these properties for the V\&V as well as
for the \emph{automated determination} of the semantics of
arbitrary LP rule inference environments is, to translate known
\textit{counter examples} into meta TCs and apply them in the
target IE. Such counter examples which show that certain semantics
do not satisfy one or more of the general properties, can be found
in literature. To demonstrate this approach we will now give some
examples derived from \cite{Dix95a,Dix95b}. For more detailed discussion of this approach and more examples see \cite{PaVV05}:\\
\begin{scriptsize}
\begin{verbatim}
Example: STABLE is not Cautious

P:  a <- neg b        P': a <- neg b
    b <- neg a            b <- neg a
    c <- neg c            c <- neg c
    c <- a                c <- a
                          c
T:{a?=>true,c?=>true}

STABLE(P) has {a, neg b, c} as its only stable model and hence it
derives 'a' and 'c', i.e. 'T' succeeds. By adding the derived atom
'c' we get another model for P' {neg a, b, c}, i.e. 'a' can no
longer derived (i.e. 'T' now fails) and cautious monotonicity is
not satisfied.

Example: STABLE does not satisfy Relevance

P:  a <- neg b           P': a <- neg b
                             c <- neg c
T:={a?=>true}

The unique stable model of 'P' is {a}. If the rule 'c <- neg c' is
added, 'a' is no longer derivable because no stable model exists.
Relevance is violated, because the truth value of 'a' depends on
atoms that are totaly unrelated with 'a'.
\end{verbatim}
\end{scriptsize}
The initial \textit{"positive"} meta TC is used to verify if the
(unknown) semantics implemented by the IE will provide the correct
answers for this particular meta test program. The "negative" TC
is then used to evaluate if the semantics of the IE satisfies the
property under tests. Such meta test sets provide us with a tool
for determining an "adequate" semantics to be used for a
particular rule-based policy/contract application. Moreover, there
are strong evidences that by taking both kinds of properties
together an arbitrary semantics might be uniquely determined by
these, i.e. via applying a meta test suite consisting of adequate
meta TCs with typical counter examples for these properties in a
IE, we can uniquely determine the semantics of this IE. Table 1
(derived from \cite{Dix95a,Dix95b}) specifies for common semantics
the properties that they satisfy.
\begin{table}[h!b!p!]
\caption{Table (General Properties of Semantics)}
\begin{tabular}{|c|c|c|c|c|c|c|c|c|c|c|}
  \hline
  Semantics & Class & Cumul. & Rat. & Taut. & GPPE & Red. & Non-Min. & Rel. & Cons. & Indep.\\
  \hline
  COMP & Normal & - & $\bullet$ & - & $\bullet$ & $\bullet$ & $\bullet$ & - & - & -  \\
  COMP$_{3}$ & Normal & $\bullet$ & $\bullet$ & - & $\bullet$ & $\bullet$ & $\bullet$ & - & - & -  \\
  WFS & Normal & $\bullet$ & $\bullet$ & $\bullet$ & $\bullet$ & $\bullet$ & $\bullet$ & $\bullet$ & $\bullet$ & $\bullet$ \\
  STABLE & Normal & - & $\bullet$ & $\bullet$ & $\bullet$ & $\bullet$ & $\bullet$ & - & - & - \\
  WGCWA & Pos. Disj. & - & $\bullet$ & - & $\bullet$ & $\bullet$ & -  & $\bullet$ & $\bullet$ & $\bullet$ \\
  CGWA & Strat. Disj. & $\bullet$ & - & $\bullet$ & $\bullet$ & $\bullet$ & $\bullet$ & $\bullet$ & $\bullet$ & $\bullet$ \\
  PERFECT & Strat.Disj. & $\bullet$ & - & $\bullet$ & $\bullet$ & $\bullet$& $\bullet$ & - & $\bullet$ & $\bullet$ \\
  \hline
\end{tabular}
\end{table}
The semantic principles described in this section are also very
important in the context of applying \textit{refactorings} to LPs.
In general, a refactoring to a rule base should optimize the rule
code without changing the semantics of the program. Removing
tautologies or non-minimal rules or applying positive/negative
reductions are typically applied in rule base refinements using
refactorings \cite{conf/ista/DietrichP05} and the semantics
equivalence relation between the original and the refined program
defined for this principles is therefore an important prerequisite
to safely apply a refactoring of this kind.

\section{Integration into Testing Frameworks and RuleML}
We have implemented the test drive approach in the ContractLog KR
\cite{PaVV05}. The ContractLog KR \cite{Paschke06} is an
expressive and efficient KR framework developed in the RBSLA
project \cite{PaschkeRBSLA} and hosted at Sourceforge for the
representation of contractual rules, policies and service level
agreements implementing several logical formalisms such as event
logics, defeasible logic, deontic logics, description logic
programs in a homogeneous LP framework as meta programs. TCs in
the ContractLog KR are homogeneously integrated into LPs and are
written in an extended ISO Prolog related scripting syntax called
Prova \cite{KPS06}. A TC script consists of (1) a unique test case
ID denoted by the function \textit{testcase(ID)}, (2) optional
input assertions such as input facts and test rules which are
added temporarily to the KB as partial modules by expressive
ID-based update functions, (3) a positive meta test rule defining
the test queries and variable bindings \textit{testSuccess(Test
Name,Optional Message for Junit)}, (4) a negative test rule
\textit{testFailure(Test Name,Message)} and (5) a \textit{runTest}
rule.

\begin{scriptsize}
\begin{verbatim}
% testcase oid
testcase("./examples/tc1.test").
% assertions via ID-based updates adding one rule and two facts
:-solve(update("tc1.test","a(X):-b(X). b(1). b(2).")).
% positive test with success message for JUnit report
testSuccess("test1","succeeded"):-
testcase(./examples/tc1.test),testQuery(a(1)).
% negative test with failure message for Junit report
testFailure("test1","can not derive a"):-
not(testSuccess("test1",Message)).
% define the active tests - used by meta program
runTest("./examples/tc1.test"):-testSuccess("test 1",Message).
\end{verbatim}
\end{scriptsize}

A TC can be temporarily loaded and removed to/from the KB for
testing purposes, using expressive ID-based update functions for
dynamic LPs \cite{PaVV05}. The TC meta program implements various
functions, e.g., to define positive and negative test queries
(\textit{testQuery, testNotQuery, testNegQuery}), expected answer
sets (variable bindings: \textit{testResults}) and quantifications
on the expected number of result (\textit{testNumberOfResults}).
It also implements the functions to compute the clause/term
specializations (\textit{specialize}) and generalizations
(\textit{generalize}) as well as the test coverage
(\textit{cover}). To proof integrity constraints we have
implemented another LP meta program in the ContractLog KR with the main test axioms:\\
\begin{scriptsize}
- $testIntegrity()$ tests the integrity of the actual program,
    i.e. it proves all integrity constrains in
    the knowledge base using them as goals constraining  on the facts and rules
    in the KB.\\
- $testIntegrity(Literal)$ tests the integrity of the literal,
i.e. it makes a hypothetical test and proves
    if the literal, which is actually not in the KB, violates any integrity constraint in the
    KB.\\
\end{scriptsize}
The first integrity test is useful to verify (test logical
integrity) and validate (test application/domain integrity) the
integrity of the actual knowledge state. The second integrity test
is useful to hypothetically test an intended  knowledge update,
e.g. test wether a conclusion from a rule (the literal denotes the
rule head) will lead to violations of the integrity of the
program. Similar sets of test axioms are provided in ContractLog
for temporarily loading, executing and unloading TCs from external
scripts at runtime.

In order to become widely accepted and useable to a broad
community of policy engineers and practitioners existing expertise
and tools in traditional SE and flexible information system (IS)
development should be adapted to the declarative test-driven
programming approach. Well-known test frameworks like JUnit
facilitate a tight integration of tests into code and allow for
automated testing and reporting in existing IDEs such as eclipse
via automated Ant tasks. The RBSLA/ ContractLog KR implements
support for JUnit based testing and test coverage reporting where
TCs can be managed in test suites (represented as LP scripts) and
automatically run by a JUnit Ant task which creates a final JUnit
and test coverage report. The RBSLA/ContractLog distribution comes
with a set of functional-, regression-, performance- and meta-TCs
for the V\&V of the inference implementations, semantics and meta
programs of the ContractLog KR w.r.t. general semantics properties
and typical adequacy criteria of KR formalisms (in particular
w.r.t. completeness, soundness, expressiveness and
efficiency/scalability).

To support distributed management and rule interchange we have
integrated TCs into RuleML (RuleML 0.9). The Rule Markup Language
(RuleML) is a standardization initiative with the goal of creating
an open, producer-independent XML/RDF based web language for
rules. The Rule Based Service Level Agreement markup language
(RBSLA) \cite{Paschke05RBSLA} which has been developed for
serialization of rule based contracts, policies and SLAs comprises
the test case layer together with several other layers extending
RuleML with modelling constructs for e.g. defeasible rules,
deontic norms, temporal event logics, reactive ECA rules. The
markup serialization syntax for test suites / test cases includes
the following constructs given in EBNF notation, i.e. alternatives
are separated by vertical bars ($|$); zero to one occurrences are
written in square brackets ($[]$) and zero to many occurrences in
braces ($\{\}$).:

\begin{scriptsize}
\begin{verbatim}
assertions ::= And test ::= Test | Query message ::= Ind | Var
TestSuite ::= [oid,] content | And TestCase ::= [oid,] {test |
Test,}, [assertions | And] Test ::= [oid,] [message | Ind | Var,]
test | Query, [answer | Substitutions] Substitutions ::= {Var, Ind
| Cterm}

Example:

<TestCase @semantics="semantics:STABLE"
class="class:Propositional">
    ...
    <Test @semantics="semantics:WFS" @label="true">
        <Ind>Test 1</Ind><Ind>Test 1 failed</Ind>
        <Query>
            <And>
                <Atom><Rel>p</Rel></Atom>
                <Naf><Atom><Rel>q</Rel></Atom></Naf>
            ...
</TestCase>
\end{verbatim}
\end{scriptsize}
The example shows a test case with the test: $test 1:
\{p=>true, not$ $q => true\}$.\\

\section{Related Work and Conclusion}
V\&V of KB systems and in particular rule based systems such as
LPs with Prolog interpreters have received much attention from the
mid '80s to the early '90s, see e.g. \cite{AHPV98}. Several V\&V
methods have been proposed, such as methods based on
\textit{operational debugging} via instrumenting the rule base and
exploring the execution trace, \textit{tabular methods}, which
pairwise compare the rules of the rule base to detect
relationships among premises and conclusions, methods based on
\textit{formal graph theory} or \textit{Petri Nets} which
translate the rules into graphs or Petri nets, methods based on
\textit{declarative debugging} which build an abstract model of
the LP and navigate through it or methods based on
\textit{algebraic interpretation} which transform a KB into an
algebraic structure, e.g. a boolean algebra which is then used to
verify the KB. As discussed in section 1 most of this approaches
are inherently complex and are not suited for the policy resp.
contract domain. Much research has also been directed at the
automated refinement of rule bases, e.g.
\cite{conf/ista/DietrichP05,BoLoRo97}, and on the automatic
generation of test cases. For an overview on rule base debugging
tools see e.g. \cite{Plant95}. There are only a few attempts
addressing test coverage measurement for test cases of
backward-reasoning rule based programs
\cite{Denney91,LBSB92,Jack96}.

Test cases for rule based policies are particular well-suited when
policies/contracts grow larger and more complex and are
maintained, possibly distributed and interchanged, by different
people. In this paper we have attempted to bridge the gap between
the test-driven techniques developed in the Software Engineering
community, on one hand, and the declarative rule based programming
approach for engineering high level policies such as SLAs, on the
other hand. We have elaborated on an approach using logic
programming as a common basis and have extended this test-driven
approach with the notion of test coverage, integrity tests,
functional and dynamic test and meta test for verify the inference
environments and their semantics properties in a open distributed
environment such as the (Semantic) Web. In addition to the
homogeneous integration of test cases into LP languages we have
introduce a markup serialization as an extension to the emerging
Semantic Web Rule Markup Language RuleML which, e.g. facilitates
rule interchange. We have implemented our approach in the
ContractLog KR \cite{Paschke06} which is based on the Prova
open-source rule environment \cite{KPS06} and applied the agile
test-driven values and practices successfully in the rule based
SLA (RBSLA) project for the development of complex, distributed
SLAs \cite{PaschkeRBSLA}. Clearly, test cases and test-driven
development is not a replacement for good programming practices
and rule code review. However, the presence of test cases helps to
safeguard the life cycle of policy/contract rules, e.g. enabling
V\&V at design/development time but also dynamic testing at
runtime. In general, the test-driven approach follows the
well-known 80-20 rule, i.e. increasing the approximation level of
the intended semantics of a rule set (a.k.a. test coverage) by
finding new adequate test cases becomes more and more difficult
with new tests delivering less and less incrementally. Hence,
under a cost-benefit perspective one has to make a break-even
point and apply a not too defensive development strategy to reach
practical levels of rule engineering and testing in larger rule
based policy or contract projects.

}

\begin{thebibliography}{99}

\bibitem{Paschke06}
A.~Paschke, M.~Bichler.
    Knowledge Representation Concepts for Automated SLA Management, Int. Journal of Decision Support Systems, to appear 2007.

\bibitem{Paschke05RBSLA}
A.~Paschke.
     RBSLA - A declarative Rule-based Service Level Agreement Language based on RuleML, Int. Conf. on Intelligent Agents, Web Technology and Internet Commerce, Vienna, Austria, 2005.

\bibitem{PaschkeRBSLA}
A.~Paschke.
    RBSLA: Rule-based Service Level Agreements. Project-Site:
    http://ibis.in.tum.de/staff/paschke/rbsla/index.htm.
    Sourceforge: https://sourceforge.net/projects/rbsla.

\bibitem{Gonz00}
A.J.~Gonzales, V.~Barr.
    Validation and verification of intelligent systems.
    {\em Journal of Experimental and Theoretical AI}.
    2000.

\bibitem{PreShi94}
A.D.~Preece and Shinghal R.
    Foundations and applications of Knowledge Base Verification.
    {\em Int. J. of Intelligent Systems}. Vol. 9, pp. 683-701, 1994.

\bibitem{CoBeCa93}
F.~Coenen and T.~Bench-Capon.
    Maintenance of Knowledge-Based Systems: Theory, Techniques and
    Tools.
    {\em Academic Press}, London, 1993.

\bibitem{AHPV98}
G.~Antoniou, F.~v.~Harmelen, R~Plant, and J~Vanthienen.
    Verification and validation of knowledge-based systems - report on two 1997 events. AI Magazine, 19(3):123–126, Fall 1998.

\bibitem{Plant95}
R.T.~Plant.
    Tools for Validation\&Verification of Knowledge-Based Systems
    1985-1995. Internet Source.

\bibitem{Preece01}
A.~Preece.
    Evaluating verification and validation methods in knowledge engineering. University of Aberdeen, 2001.

\bibitem{BOP06}
    P.~Bonatti, D.~Olmedilla, and J~Peer. Advanced
policy explanations. In 17th European Conference on Artificial
Intelligence (ECAI 2006), Riva del Garda, Italy, Aug-Sep 2006. IOS
Press.

\bibitem{KPS06}
    A.~Kozlenkov, A.~Paschke, M.~Schroeder, Prova - A Language for Rule Based Java Scripting, Information Integration, and Agent Programming.
http://prova.ws., 2006.

\bibitem{Llyod87}
  J.W. Lloyd. Foundations of Logic Programming. 1987, Berlin: Springer.

\bibitem{DIX95}
J.~ Dix.
    Semantics of Logic Programs: Their Intuitions and Formal Properties. An Overview. In Andre Fuhrmann and Hans Rott, editors, Logic, Action and Information -- Essays on Logic in Philosophy and Artificial Intelligence, pages 241--327. DeGruyter, 1995.

\bibitem{Dix95a}
J.~Dix.
    A Classification-Theory of Semantics of Normal Logic Programs: I. Strong Properties," Fundamenta Informaticae XXII(3) pp. 227-255, 1995.

\bibitem{Dix95b}
J.~Dix.
    A Classification-Theory of Semantics of Normal Logic Programs: II. Weak Properties. Fundamenta Informaticae, XXII(3):257-288, 1995.

\bibitem{PLOTKIN70}
G.D. Plotkin.
  A note on inductive generalization.
  {\em Machine Intelligence}, 5, 1970.

\bibitem{PaDiBo05}
A.~Paschke, J. Dietrich and H.~Boley.
    W3C RIF Use Case: Rule Interchange Through Test-Driven Verification and
    Validation. http://www.w3.org/2005/rules/wg/wiki/ Rule\_Interchange\_Through\_Test-Driven\_Verification\_and\_Validation,
    2005.

\bibitem{DEGV97}
E.~Dantsin, T.~Eiter, G.~Gottlob, and A.~Voronkov.
     Complexity and expressive power of logic programming. IEEE Conference on Computational Complexity, pages 82--101, Ulm, Germany, 1997.

\bibitem{conf/ista/DietrichP05}
J.~Dietrich and A.~Paschke.
  On the test-driven development and validation of business rules.
  Int. Conf. ISTA'2005, 23-25 May, 2005, Palmerston North, New Zealand, 2005.

\bibitem{PaVV05}
A. Paschke.
     The ContractLog Approach Towards Test-driven Verification and Validation of Rule Bases - A Homogeneous
Integration of Test Cases and Integrity Constraints into Dynamic
Logic Programs and Rule Markup Languages (RuleML), IBIS, TUM,
Technical Report 10/05.

\bibitem{KLM90}
S. Kraus, D. Lehmann, M. Magidor.
  Nonmonotonic reasoning, preferential models and cumulative logics.
  {\em Artificial Intelligence}, 44(1-2):167--207, 1990.

\bibitem{BoLoRo97}
F.~Bouali, S.~Loiseau,M-C.~Rousset.
    Verification and Revision of Rule Bases.
    {\em in: J.~Hunt,R.~Miles(Eds.)}, Reserach and Development in
    Expert System, SGES publication, pp. 253-264.

\bibitem{Mes93}
P.~Meseguer.
    Expert System Validation Through Knowledge Base Refinement.
    {\em IJCAI'93}, 1993.

\bibitem{ChCoSt90}
C.L.~Chang, J.B.~Combs, R.A.~Stachowitz.
    A Report on the Expert Systems Validation Associate (EVA).
    {\em Expert Systems with Applications}, Vol.1,No.3,pp. 219-230.

\bibitem{Denney91}
R.~Denney.
 Test-Case Generation from Prolog-Based
Specifications, IEEE Software, vol. 8,  no. 2,  pp. 49-57,
Mar/Apr,  1991.

\bibitem{Jack96}
O.~Jack.
     Software Testing for Conventional and Logic Programming,
vol. 10 of Programming Complex Systems, F. Belli (ed.), Walter de
Gruyter \& Co., Berlin, New York, 1996.


\bibitem{LBSB92}
 G. Luo, G. Bochmann, B. Sarikaya, and M. Boyer.
 Control-flow based testing of prolog programs. In Int. Symp. on Software
Reliability Enginnering, pp. 104-113, 1992.

\bibitem{Zan95} Zaniolo, C. Active Database Rules with Transaction-Conscious Stable-Model Semantics. in Int. Conf. on Deductive and Object-Oriented Databases. 1995.
\bibitem{FlGr98} Flesca, S. and S. Greco. Declarative Semantics for Active Rules. in Int. Conf. on Database and Expert Systems Applications. 1998: Springer LNCS.
\bibitem{BaLo96} Baral, C. and J. Lobo. Formal characterization of active databases. in Int. Workshop on Logic in Databases. 1996.
\bibitem{Luda98}    Ludascher, B., Integration of Active and Deductive Database Rules, Phd thesis,, in Institut für Informatik. 1998, Universität Freiburg, Germany.
\bibitem{LaLuMa98} Lausen, G., B. Ludascher, and W. May, On Logical Foundations of Active Databases. Logics for Databases and Information Systems, 1998: p. 389-422.
\bibitem{Zan93} Zaniolo, C. A unified semantics for Active Databases. in Int. Workshop on Rules in Database Systems. 1993. Edinburgh, U.K
\bibitem{BonKi95} Bonner, A.J. and M. Kifer, Transaction logic programming (or a logic of declarative and procedural knowledge). 1995, University of Toronto.
\bibitem{Rasch94} Raschid, L.: A semantics for a class of stratified production system programs. Int. Journal of Logic Programming, 21(1), pp. 31-57, 1994.
\bibitem{DuMan02} Dung, M.P., Mancarella, P. Production Systems with Negation as Failure, IEEE Transactions on Knowledge and Data Engineering, V14(2), 2002.
\bibitem{ECA-RuleML} Paschke, A., ECA-RuleML: An Approach combining ECA Rules with interval-based Event
  Logics,Internet-based Information Systems, Technical Report 11/2005, Technical University Munich, 2005, November,
  http://ibis.in.tum.de/staff/paschke/rbsla/
\bibitem{ECA-LP}
    Paschke, A., ECA-LP: A Homogeneous Event-Condition-Action Logic Programming Language,
    Internet-based Information Systems, Technical Report 11/2005, Technical University Munich, 2005, November,
    http://ibis.in.tum.de/staff/paschke/rbsla/.

\end{thebibliography}
\end{document}